\documentclass[conference]{IEEEtran}
\IEEEoverridecommandlockouts
\usepackage{cite}
\usepackage{amsmath,amssymb,amsfonts}
\usepackage{enumitem}
\usepackage{algorithmic}
\usepackage{graphicx}
\usepackage{textcomp}
\usepackage{url}
\usepackage{xcolor}
\def\BibTeX{{\rm B\kern-.05em{\sc i\kern-.025em b}\kern-.08em
    T\kern-.1667em\lower.7ex\hbox{E}\kern-.125emX}}

\definecolor{PrologPredicate}{RGB}{0,0,200}
\definecolor{PrologVar}      {RGB}{145,032,039}
\definecolor{PrologComment}  {RGB}{169,082,044}
\definecolor{PrologOther}    {rgb}{0.2,0.2,0.2}
\definecolor{PrologString}   {RGB}{070,120,200}
\usepackage{relsize}
\usepackage{listings} 

\lstdefinestyle{MyInline}
{
  basicstyle = \ttfamily\color{PrologOther},
  breaklines = true,
  breakatwhitespace=true,
  upquote = true,
}
\lstdefinestyle{MySCASP}
{
  keywords = {},
  breaklines = true,
  upquote = true,
  basicstyle = \relsize{-.5}\ttfamily\color{PrologPredicate},
  moredelim = {*[s][\color{black!40!PrologPredicate}]{\#pred}{.}},
  moredelim = {*[s][\color{black!40!PrologPredicate}]{\#show}{.}},
  moredelim = {*[s][\color{black!40!PrologPredicate}]{\#hide}{.}},
  moredelim = {*[s][\color{PrologVar}]{(}{)}},
  moredelim = {*[s][\color{PrologString}]{'}{'}},
  moredelim = {*[s][\color{PrologOther}]{:-}{.}},
  commentstyle = \mdseries\color{PrologComment},
  morecomment=[l]\%,
}
\begin{document}

\title{Reliable Collaborative Conversational Agent System\\based on LLMs and Answer Set Programming}

\author{\IEEEauthorblockN{1\textsuperscript{st} Yankai Zeng}
\IEEEauthorblockA{\textit{Department of Computer Science} \\
\textit{University of Texas at Dallas}\\
Richardson, USA \\
yankai.zeng@utdallas.edu}
\and
\IEEEauthorblockN{2\textsuperscript{nd} Gopal Gupta}
\IEEEauthorblockA{\textit{Department of Computer Science} \\
\textit{University of Texas at Dallas}\\
Richardson, USA \\
gupta@utdallas.edu}
}

\maketitle

\begin{abstract}
As the Large-Language-Model-driven (LLM-driven) Artificial Intelligence (AI) bots became popular, people realized their strong potential in Task-Oriented Dialogue (TOD). However, bots relying wholly on LLMs are unreliable in their knowledge, and whether they can finally produce a correct outcome for the task is not guaranteed. The collaboration among these agents also remains a challenge, since the necessary information to convey is unclear, and the information transfer is by prompts---unreliable, and malicious knowledge is easy to inject.
With the help of knowledge representation and reasoning tools such as Answer Set Programming (ASP), conversational agents can be built safely and reliably, and communication among the agents made more reliable as well. We propose an Administrator-Assistant Dual-Agent paradigm, where ASP-driven bots share the same knowledge base and complete their assigned tasks independently. The agents communicate with each other through the knowledgebase ensuring consistency.
The knowledge and information conveyed are encapsulated and invisible to the users, ensuring the security of information transmission. To illustrate the dual-agent conversational agent paradigm, we have constructed AutoManager, a dual-agent system for managing the drive-through window of a fast-food restaurant such as Taco Bell in the US. In AutoManager, the assistant bot takes the customer's order while the administrator bot manages the menu and food supply. We evaluated our AutoManager system and compared it with the real-world Taco Bell Drive-Thru AI Order Taker, and the results show that our method is more reliable.
\end{abstract}

\begin{IEEEkeywords}
Answer Set Programming, s(CASP), Large Language Model, chatbot
\end{IEEEkeywords}

\section{Introduction}

Collaboration is ubiquitous in human society and can even be regarded as the very foundation upon which societies and civilizations are built. Enterprises, for example, require diverse roles to fulfill specialized functions---marketing, technology, finance, legal affairs, etc.---and only through collaboration among these distinct positions can a company operate effectively. The same principle applies to conversational agents. Different agents serve varied user groups and accomplish distinct tasks, while their inter-agent collaboration sustains the harmonious operation of the entire system.

For instance, consider automating an airport check-in operation with AI agents. An Operations Management bot could allocate boarding gates and respond to flight delays or emergencies, while a Counter Assistant bot assists passengers with check-in, seat selection, baggage check-in, class upgrades, and real-time information dissemination. Things may, however, change dynamically on the ground, e.g., an airport gate suffers a mechanical fault and the aircraft has to be re-assigned to another gate. In such scenarios, the real-time and accurate information exchange between agents and the dynamic and contextually appropriate responses to received information demands heightened attention.

Collaboration between such AI bots requires information known to one bot to be made instantaneously available to the other. For conversational bots solely based on large language models (LLMs), the instantaneous information can only be passed through prompts and cannot be obtained through training or fine-tuning (though  Therefore,  instantaneous information will occupy a large amount of prompt space, thereby reducing the operating efficiency and reliability of the LLM. In addition, due to the unreliable reasoning and hallucinatory nature of LLMs, the instantaneous information cannot always be used properly to complete the tasks.
%
Therefore, task-oriented bots that are based on LLMs have not always succeeded. A recent unsuccessful attempt is reported by McDonald \cite{failed-mcdonald-bot} for developing drive-through window order-taking chatbots. A common issue arose from the absence of a reliable communication mechanism: the AI order taker fails to receive real-time updates on ingredient availability in restaurants, for example. Even when such information is accurately transmitted, the chatbot may neglect or lack sufficient reasoning capacity to initiate appropriate actions, such as notifying customers of shortages, resulting in the inability to fill an order, for example. Another AI order taker, delployed by Taco Bell \cite{tacobell-bot}, also needs considerable customized training along with constant supervision by a human being. Therefore, they are essentially assistants to humans rather than systems that can operate independently. 


Motivated by the above problem, we propose the development of \textbf{reliable} knowledge-driven collaborative chatbots. 
Different from other agentic AI systems, our method focuses on the \textbf{accuracy} and \textbf{controllability} of the agent-wide communication, and employs the reasoning system to make the communication \textbf{reliable}.
We limit ourselves to the dual-agent collaboration in this paper, i.e., only two chatbots working collaboratively are considered. Typically, such a dual-agent system consists of a manager bot to whom changes brought about by instantaneous events are communicated by the human manager, and a customer service chatbot that helps human users (human customers). 

In our earlier work, we developed task-oriented chatbots \cite{autoconcierge} as well as socialbots \cite{autocompanion} that are knowledge-driven and thus reliable. To ensure the reliability of chatbot responses, we limited the use of LLMs as a semantic parser. We used answer set programming (ASP), a knowledge representation and reasoning formalism, to perform the backend reasoning. The reasoning that we perform is true deductive or abductive reasoning and is distinct from ``reasoning" performed by Large Reasoning Models. The LLM-based semantic parser turns text into knowledge represented as logic predicates. ASP is then used for performing reasoning over these predicates. Pre-existing commonsense knowledge represented in ASP is also employed during this reasoning. The computed response is a set of predicates that are then converted into text, using an LLM as a reverse logic predicates to English text translator. 
Each chatbot in the collaboration system is constructed following the above method. Each single bot performs a distinct but complementary task by interacting with a specific type of human user, and communicates with the other through a shared knowledge base.

%
%
%
%
Based on the dual-agent collaboration system, we built AutoManager to model the collaboration for running a Taco Bell restaurant, where the manager bot manages the food supply and menu updates, and the customer service bot takes customer orders. The customer service bot chats with the customers to directly take orders, while the manager bot chats with an actual human manager who provides instructions in natural language to keep the shared knowledge consistent. We evaluated the conversation quality in the AutoManager manually and compared it with the real-world Taco Bell Drive-Thru AI system, which is largely based on machine learning technology. The result shows that our AutoManager, based on a hybrid of LLM and ASP, is more reliable than the real-world Taco Bell AI.

The main contribution of our paper is the KRR-driven multi-agent collaborating chatbot paradigm and its use in developing a practical reliable application that was hitherto not possible. 
%


\section{Background}

\subsection{Large Language Models (LLMs):}
Until recently, transformer-based Large Language Models, pre-trained on an enormous quantity of well-annotated data, have been applied to general NLP tasks. With the advent of Large Language Models, the paradigm changed from pre-training and fine-tuning (by \cite{pre_trained_transformers}) to teaching a language model any arbitrary task using just a few demonstrations, called \textit{in-context learning}, or \textit{prompt engineering}. \cite{gpt3} introduced an LLM called GPT-3 containing approximately 175 billion parameters that have been trained on a massive corpus of filtered online text, on which the well-known ChatGPT is based. 
GPT-3, as well as the later GPT-4 (by \cite{gpt4}), can perform competitively on several tasks such as question-answering, semantic parsing, and machine translation. However, such LLMs lack the ability of precise logical reasoning and find it hard to overcome the hallucination brought about by the training data, according to \cite{gpt3-scope,chain,chatgpt-critic}. 

\subsection{Answer Set Programming and s(CASP)}
%
Answer Set Programming (ASP) \cite{cacm-asp,gupta-csr} is a logic programming paradigm suited for knowledge representation and reasoning that facilitates commonsense reasoning.  
The s(CASP) system \cite{scasp} is an answer set programming system that supports predicates, constraints over non-ground variables, uninterpreted functions, and, most importantly, a top-down, query-driven execution strategy.
These features make it possible to return answers with non-ground variables (possibly including constraints among them) and compute partial models by returning only the fragment of a stable model necessary to support the answer.
%

Complex commonsense knowledge can be represented in ASP, and the s(CASP) query-driven predicate ASP system can be used for querying it. Commonsense knowledge can be emulated using (i) default rules, (ii) integrity constraints, and (iii) multiple possible worlds by \cite{gelfondkahl,gupta-csr}. Default rules are used for jumping to a conclusion in the absence of exceptions, e.g., a bird normally flies, unless it's a penguin. 


\begin{lstlisting}[style=MySCASP]
flies(X) :- bird(X), not abnormal_bird(X).
abnormal_bird(X) :- penguin(X).
\end{lstlisting}

%

\noindent Integrity constraints allow us to express impossible situations and invariants. For example, a person cannot sit and stand at the same time.

\begin{lstlisting}[style=MySCASP]
false :- person(X), sit(X), stand(X).
\end{lstlisting}

\noindent Finally, multiple possible worlds allow us to construct alternative universes that may have some parts in common but other parts inconsistent. For example, the cartoon world of children's books has a lot in common with the real world (e.g., birds can fly in both worlds), yet in the former, birds can talk like humans, but in the latter, they cannot. 
Default rules are used to model a bulk of our commonsense knowledge. Integrity constraints help in checking the consistency of the information extracted. Multiple possible worlds allow us to perform assumption-based (or abductive) reasoning. 

A large number of commonsense reasoning applications have already been developed using ASP and the s(CASP) system by \cite{blawx,logical-english,chef,murder-trial}. 
The query-driven s(CASP) ASP system was crucial for the AutoConcierge system developed by \cite{autoconcierge} earlier. s(CASP) was used to perform commonsense reasoning resembling a human concierge: (i) To check for consistency and completeness in the knowledge derived from the user's dialogs, and to ask further questions and seek more information in case it was inconsistent or incomplete. (ii) Subsequently, concluding the extracted knowledge. Explanation for each response can also be given by justifications for successful queries as proof trees, according to \cite{scasp-justification}.  \cite{autocompanion} also developed a Socialbot that can hold conversations with a stranger about movies, books, and sports using similar techniques.

\subsection{The STAR Framework}

The reliable reasoning chatbots are built by emulating humans thinking. When humans hear a sentence, they parse it to extract its meaning and represent the meaning in their minds as knowledge. Humans then check for consistency and correctness of this knowledge using additional (commonsense) knowledge that also resides in their minds. Next, humans will find gaps in this acquired knowledge and attempt to fill it by asking further questions.  

Therefore, similar to humans, when the sentences from the user come to the chatbot, it will also process them in three stages:
\textbf{First}, convert the natural language input from the human user to knowledge (represented as pre-determined logic predicates), which is done by LLMs outstandingly well.
\textbf{Next}, check the consistency and correctness of the input knowledge and use the knowledge extracted from the dialog to reason about the next action. The ASP reasoning system ensures this capability.
\textbf{Finally}, the knowledge representing the next step, computed by the ASP engine, is converted into a natural language and communicated by another LLM back to the human user. This cycle continues. 

This mechanism is embodied in the Semantic-parsing Transformer and ASP Reasoner (STAR) framework, proposed by \cite{star}, leveraging the advantages of LLMs and ASP systems by combining them systematically. It parses the semantics of the text sentences to generate the predicates using LLMs such as those from the GPT series. After that, it sends the predicates to the ASP system to get reliable answers through reasoning. This framework was previously applied to task-oriented chatbots and socialbots, such as AutoConcierge by \cite{autoconcierge} and AutoCompanion by \cite{autocompanion}, and has proved to work well.

\section{Dual-Agent Chatbot Design}
As the name indicates, the dual-agent collaborative system consists of two conversational agents who share the knowledge base but have different functions. In such a dual system, the manager bot manages and maintains the knowledge base information, while the customer service bot offers services to customers using the knowledge base. 
Typically, in a restaurant, for example, the manager bot maintains the menu and food supplies, and the customer service bot takes orders and answers the customers' questions. 

\begin{figure}[ht]
    \centering
    \includegraphics[width=0.85\linewidth]{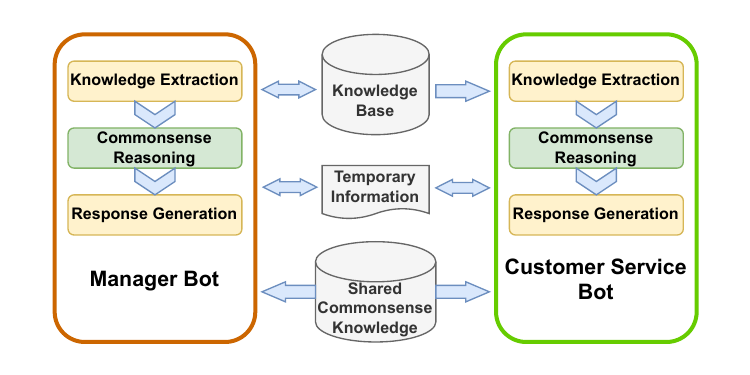}
    \caption{AutoManager Architecture. The manager and customer service bot are constructed in the STAR framework. Two bots share the knowledge base, temporary information list, and a collaborative rule set.}
    \label{fig:dual_arch}
\vspace{-0.15in}
\end{figure}

Both bots are constructed based on the STAR framework, in which the LLM is only used for parsing natural language into a pre-defined predicate logic-based vocabulary and vice versa. All the decision-making and consistency-checking steps are done by a reasoning engine driven by the s(CASP) goal-directed ASP system from \cite{scasp}. This framework ensures the reliability of the decision and the robustness of the input and output. The administrator and assistant bots' functions in AutoManager are shown in Figure \ref{fig:dual_arch} and will be detailed below.

\subsection{Manager Bot Functions}
In AutoManager, the manager bot maintains the menu and food supplements. The users (who represent restaurant employees and managers) can change the menu by adding, deleting, or modifying the menu items, and notifying of the shortage or restoration of some ingredients. In this work, we use ``dish'' to represent any item that can be ordered on the menu except combos, ingredients, and sauces. On the other hand, the term ``food'' refers to the general name for the combo, dish, ingredient, and sauce. 

\smallskip\noindent\textbf{LLM Parser and Ontology}.
In AutoManager, both the manager and customer service bot leverage an LLM (GPT-4) for extracting knowledge from a manager's and customer's dialog respectively. This knowledge is represented as predicates, that come from a pre-determined vocabulary. Per the STAR framework, this logical vocabulary can be thought of as a language that the bot ``understands". The LLM is used primarily to translate natural language to logic predicates and vice versa. The predicates corresponding to a natural language sentence can be thought of as representing its meaning or linguistic \textit{deep structure}.

To capture the main functions of the manager bot, the following predicates have been pre-defined: \textit{runout}, \textit{restore}, \textit{add}, \textit{edit}, and \textit{delete}. These predicates have distinct meanings wrt the commonsense reasoning rules defined for the administrator bot. The predicate \textit{runout/1} and \textit{restore/1} indicate that a certain ingredient is out of stock or has been restored. The predicate \textit{add(Type, Food)} means to add a kind of food to the menu, where the first arity is the type to add, and the second arity is the name of the food. This kind of food can either be a dish (e.g., Grilled Beef Taco), a combo of dishes (e.g., Grilled Beef Taco Combo), or an ingredient (e.g., Grilled Beef). If the user does not specify the type of food, the LLM parser will extract \textit{add(Food)} instead. Otherwise, ``Type'' can be either ``dish'', ``combo'', or ``ingredient''. If the user wants to delete a food from the menu, the predicate \textit{delete(Food)} is what they need. The predicates \textit{add(Food, Property, Value)}, \textit{edit(Food, Property, Value)}, \textit{edit(Food, Property, Original\_Value, New\_Value)}, \textit{delete(Food, Property, Value)}, and \textit{delete(Food, Property)} express the users' intentions of modifying the property of some food to some value. For example, if the restaurant runner wants to change the price of the Grilled Cheese Burrito to \$3.80, the predicate generated would be:
{\small \tt edit(`Grilled Cheese Burrito', price, 3.80).}
And if the runner wants to change one of the ingredients of Crunchy Taco from cheese to lettuce, the predicate
{\small \tt edit(`Crunchy Taco', included\_ingredient, `Cheese', `Lettuce').}
will be generated. Also, predicate \textit{quit/0} and \textit{irrelevant/0} show that the user wants to leave, or said something irrelevant. These predicates, along with their usage, are sent as prompts to the LLM, where the prompt also includes a few simple examples to define the format that the LLM will follow.

\smallskip\noindent\textbf{Conversational Knowledge Template (CKT)}. CKTs, proposed by \cite{ckt}, are state machines that represent the logic that a bot must follow to generate a response. The CKTs are coded in ASP. One characteristic of human input is that it can contain variable information. For example, a user may say ``I want to order a soft taco,'' and then say ``Can you please replace meat with beans?" or a user may say ``I want to order a soft taco with beans instead of meat." CKTs can handle a variable amount of information contained in each sentence. 
In the manager bot, to add a type of food, CKT will loop in asking for the information listed below: the category of the food, its price, its ingredients, the calories it contains, and the popular toppings to recommend. As for adding a combo, CKT will ask for its price, calories contained, and the food it includes. Once the CKT decides what to ask for in the next round of dialogue, a predicate \textit{ask(Name, Property)} will be generated. Note that CKTs are flexible structures. The natural language input can be provided by a human in one shot or piecemeal over multiple responses. 

\smallskip\noindent\textbf{Response Generation}.
The response generated by the manager bot includes two parts: confirmation and question. The predicates for confirmation are directly correspond to the user's input. For instance, when the user says that the lettuce is out of stock, predicate \textit{confirm(`run out', `Lettuce')} will be applied as the confirmation, which will later be sent to an LLM for generating a natural language sentence like \textit{``We are short of lettuce, understood.''} The question part is only for the case where a CKT is invoked, e.g., adding a new dish to the menu. Similar to the predicate extraction part, we provide the LLM with the usage of each predicate and several examples in the prompt. We also add instructions to the prompt to tell the LLM the background of the dialogue; consequently, the LLM can generate suitable sentences as responses.

\subsection{Customer Service Bot Functions}
The customer service bot of AutoManager is responsible for users' orders. Its main functions include helping users complete their orders, answering their questions, and recommending foods according to their preferences. The customer service bot's logic is much more complex than a manager bot's because food ordering instructions at a fast food drive-through can be complex.
 
\smallskip\noindent\textbf{LLM Parser and Ontology}.
Like the manager bot, the customer service bot uses an LLM with instructions, predicate usages, and examples as prompts for knowledge extraction. In the customer service bot, the LLM extracts six predicates: \textit{need\_recommend/2}, \textit{order/2}, \textit{specify/2}, \textit{update/3}, \textit{complete/0}, and \textit{query/2}.
The predicate \textit{need\_recommend(Content, Type)} captures the user's intention that asks for a recommendation. The `Type' arity can be either `category' or `upgrade'. Suppose the `Type' is `category'. In that case, it means that the user needs the recommendation of a specific category of food, and then the `Content' arity can be either `taco', `burrito', `quesadilla', `nacho', `bowl', `vegan', `chicken', etc., or even `all', indicating all categories of foods. If `Type' is `upgrade', indicating the user needs a recommendation for the popular toppings for the dish, then the `Content' will be the name of the dish, such as `Crunchy Taco', etc.
The predicate \textit{order(Food, Number)} captures the user's intention of ordering, where the `Food' arity is the name of a dish or combo that the user wants to order. The `Number' arity indicates the quantity the user orders. The predicate \textit{specify(Combo, Dish)} specifies a certain dish as the combo member, such as designating a specific taco for the taco bundle, or a drink for some combo. This predicate is used only for the combo. The predicate \textit{update(Dish, Operation, Option)} captures the user's intention of updating an ordered dish. The arity `Dish' refers to the dish name, the `Operation' arity refers to the upgrading operations, and the `Option' refers to the corresponding options to choose from. In operations `change', `add', `no', `less', and `extra', the option tells what ingredient to modify. For the operations `fresco', `supreme', and `grill', the option can only be `yes' or `no', showing whether the user would like these upgrades. For the operation `size', the option can be `regular' or `large'.
The predicate \textit{query(Category, Food)} captures the user's intention of asking. The `Category' arity refers to the type of information the user asks for, while the `Food' arity shows the object the user refers to. The `Food' can also be `all' in some cases. For example, the question ``What are the available toppings?'' can be expressed by the predicate
{\tt \small query(`add-on', `all')},
and the question ``What's the price of the Veggie Mexican Pizza?'' can be expressed by the predicate
{\tt \small query(`price', `Veggie Mexican Pizza')}. When the user says that they have finished all they need to order, the predicate \textit{completed} will be extracted.

\smallskip\noindent\textbf{Ordering Strategy}.
After extracting predicates from the users' input, the ASP-based goal-directed reasoning system decides what to say in the next round of dialogue. For the first several rounds of conversation, the customer service bot will listen to the customer's requirements, and only reply by saying ``I got it, what else do you want to order?'' unless the user wants the bot to recommend food or answer questions. Once the user has indicated that he or she has completed all the dishes he or she wishes to order, where the predicate \textit{completed} is extracted, the customer service bot, driven by CKTs, will check if any important information is missing and ask the customer corresponding questions. If the customer orders a single dish, the service bot will ask if they would like some toppings on the dish, and for the drink, the service bot will ask for the size, if not specified. If the order contains any combo, the CKT will go through each dish in the combo, asking the user to choose one optional dish from the listed choices, and then complete the CKT loop for this dish. Since the user may order the same dish repeatedly, the CKT will execute the same loop several times until all the dishes are complete. In this ``ask'' mode, the next-action predicate \textit{ask(Food, Option)} or \textit{ask(Combo, Food, Option)} is generated.

\smallskip\noindent\textbf{Price Calculation}.
After collecting all the information, the service bot will calculate the price of the order. An extra charge will be applied if the order contains upgrades or add-ons. The output predicates, categorized as ``check'' mode, contain all the ordered dishes with their special requirements, grouped by each ordered dish, and the total price, as the example below shows. 

\begin{lstlisting}[style=MySCASP]
[order(`Combo A'), specify(`Crunchy Taco'), specify(`Soft Taco'), add(`Purple Cabbage'), specify(`Pepsi'), size(`large')], [order(`Combo A'), specify(`Soft Taco'), make_supreme(`yes'), specify(`Nacho Cheese Doritos Locos Tacos'), extra(`Beans'), specify(`Pepsi'), size(`regular')], price(12.68).
\end{lstlisting}

\smallskip\noindent\textbf{Question Answering and Recommendation}.
The question-answering function will be triggered if a \textit{query/2} predicate is extracted, and the customer service bot will look up an answer in the knowledge base. If the \textit{need\_recommend/2} predicate is captured from the user input, the service bot will recommend dishes that satisfy the user's requirements. Those mentioned in the previous conversation will not be recommended again. The ``answer'' mode produces the next-action predicate \textit{answer(Food, Property, Value)} for question answering, and the ``recommend'' mode generates \textit{recommend(Type, Food)} predicate to help the users find the food they would like.

\smallskip\noindent\textbf{Response Generation}.
The response is generated using an LLM with some templates and prompts. In the ``quit'' mode, the only predicate output from the reasoning engine is \textit{quit/0}, and the LLM will generate the response by rewriting the template for saying goodbye. Similarly, in the ``else'' mode, the LLM will respond by asking what else to help with. For the other modes, including ``ask'', ``recommend'', ``answer'', and ``check'', the response predicates contain a confirmation and the action part. The confirmation makes sure the LLM extracts the user's intention correctly. For example, the extracted predicate
{\tt \small update(`Soft Taco', `add', `Cheese').}
leads to the confirmation predicate
{\tt\small confirm(`add', `Cheese').},
and generates the response like ``Okay, you'd like to add more cheese.'' in the LLM response generation step. The confirmation to the \textit{irrelevant} predicate gives a response like ``Sorry, I don't understand. Let's get back to your order.'', while the confirmation for unavailable dishes will be explained in Section \ref{sec:unav}. The output of the action part of each mode includes different predicates depending on the action mode, which is described in the above paragraphs. All the output predicates will be sent to an LLM for the natural language response generation. Like the knowledge extraction part, the prompt here also contains the meaning and usage of each predicate and a few examples.

\section{Dual-Agent Collaboration}
\label{sec:unav}
Our AutoManager Chatbot system is designed as a manager-customer-service system managing a Taco Bell restaurant. The manager bot and the customer service bot collaborate on food sales and inventory management, where the manager bot maintains the inventory and the service bot interactively helps customers place their orders accordingly. 
To maintain the system to work effectively, the two agents should achieve the following functions:
(i) Shareability: The two bots share the same knowledge base and can make changes to it if necessary. In AutoManager, the manager and service bot share the same menu, and the manager bot can update it, such as adding new dishes or adjusting the price. The menu is part of the knowledge base. 
(ii) Independence: Without the other agent, each bot alone should still run properly, even if the other bot is dormant, since each bot is designed for solving an independent task, and the only difference between the two bots is their target user. For example, in AutoManager, the manager phase is only for restaurant employees, while the customer service phase interacts with restaurant customers.
(iii) Collaboration: However, despite being independent, the two bots still constitute a dual system, meaning that they can communicate with each other through the knowledge base, which holds the key information for both bots to complete their respective tasks. In AutoManager, if the storage of lettuce, for instance, is low, the manager bot will communicate this information to the customer service bot by updating the knowledge base, and the service bot will then reject the orders containing lettuce. 

\smallskip\noindent\textbf{Shared Commonsense Knowledge and Rules:}
In AutoManager, the information about the availability of ingredients (expressed by predicates \textit{runout/1} and \textit{restore/1}) is captured from the dialogue by the manager bot and saved in a temporary state file for the customer service bot to read. Apart from these predicates, some rules that reason on these predicates are also necessary for both bots: the order-taking bot would want to know what dishes are currently unavailable from the manager bot and communicate this to the customer. Both cases require the rules regarding the shortage of ingredients for the unavailable dishes to be explained, shared by the collaborative rule set. The shared rules in AutoManager are listed below.

\begin{lstlisting}[style=MySCASP]
unavailable(Dish) :- dish(Dish), include_ingredient(Dish, Ingredient), runout(Ingredient).
unavailable(Ingredient) :- ingredient(Ingredient), runout(Ingredient).
unavailable(Sauce) :- sauce(Sauce), runout(Sauce).
unavailable(Combo) :- combo(Combo), combo_contain(Combo, Dish), unavailable(Dish).
unavailable(Combo) :- combo(Combo), combo_contain(Combo, Group_Name), combo_option\_group(Group_Name, L), all_unavailable(L).

\end{lstlisting}

\smallskip\noindent\textbf{Leveraging External Information For Internal Tasks (LEIFIT) Mechanism:}
With commonsense knowledge and rules regarding unavailability, the service bot can use the predicate \textit{unavailable/1} to complete the ordering task. This direct use of shared predicates and rules for reasoning is called the LEIFIT mechanism. In Automanager, for example, this \textit{unavailable/1} predicate is used for rejecting user requirements that cannot be fulfilled and listing proper answers to the user's questions and recommendation demands. For each round of conversation, the reasoning engine first checks if the dish or topping the user orders is available according to the shared rules. If there is more than one unavailable order, the bot will generate a list of the unavailable dishes and the missing ingredient as the confirmation, e.g., 

\begin{lstlisting}[style=MySCASP]
confirm(`unavailable', [unavailable(`Cantina Chicken Burrito', runout(`Slow-Roasted Chicken')), unavailable(`Cantina Chicken Taco', runout(`Slow-Roasted Chicken'))])
\end{lstlisting}

\noindent Meanwhile, this user requirement will not be updated on the order list. In the ``recommend'' mode, the reasoning engine will exclude those unavailable dishes from the recommendation; in the ``answer'' mode, when the user asks for the available toppings of some dishes, those in short supply will not be included.

In the LEIFIT mechanism, we ensure each round of conversation updates the temporary information list once and checks the consistency of the shared commonsense knowledge once for security reasons. The update on the knowledge base will only be executed after both bots end their conversations. The update operation is considered an atomic operation to avoid inconsistency during the update.

\section{Implementation}
AutoManager uses Python to connect the LLM-based text-to-knowledge and knowledge-to-text translation system with the ASP-based reasoning system. We use s(CASP) goal-directed ASP system as the reasoning engine. GPT-4 is chosen as the LLM in both the LLM Parser and Response Generation stages. 

\smallskip\noindent\textbf{Data Collection:}
AutoManager is built to simulate a Taco Bell restaurant. Therefore, the menu is constructed based on complete Taco Bell's menu \footnote{\url{https://www.tacobell.com/food}}. Every single dish is expressed by the predicate \textit{dish/1}, while the combos, ingredients, and sauces are recorded by the predicates \textit{combo/1}, \textit{ingredient/1}, and \textit{sauce/1}. The predicate \textit{original\_price(Food, Price)} shows the original price of each food before the personal change, and the predicate \textit{original\_cal(Food, Calories)} represents the original calories the food carries. For a dish, the predicate \textit{category(Food, Type)} classifies it as a taco, a burrito, or some other type. The knowledge base uses predicates \textit{included\_ingredient(Dish, Ingredient)}, \textit{replaceable\_ingredient(Dish, Original, Replacement)}, 
\textit{replacement\_price(Dish, Original, Replacement, Price)}, \textit{available\_topping(Dish, Topping)}, \textit{popular\_topping(Dish, Topping)}, \textit{upgrade\_price(Dish, Topping, Price)}, \textit{upgrade\_cal(Dish, Topping, Calories)}, \textit{available\_special\_style(Dish, Style)}, and \textit{special\_style\_price(Dish, Style, Price)}
, etc., to represent all information for a single dish.
, where the special style could be either `fresco', `supreme', or `grill'. The predicates \textit{extra\_price/2} and \textit{extra\_cal/2} are used to capture information on the extra size of each topping. The combo-related predicates include \textit{combo\_contain(Combo, Dish)}. Here, the `Food' arity in the predicate \textit{combo\_contain/2} can be either a dish or a group name, such as side, drink, taco, etc., which can be expressed by the predicate \textit{combo\_option\_group(Group\_Name, List)}. Within a group, the dishes can have a price difference. Therefore, we use the predicate \textit{group\_upgrade\_price(Group\_Name, Dish, Price)} to represent those who need an extra charge. The predicate \textit{size\_changable\_drink/1} records the drinks that the customer can upgrade to `large' size, and the predicate \textit{upgrade\_size\_price/1} records the unified upgrade price. 
We also record some special categories of food using the predicate \textit{veggie/1}, \textit{cantina\_chicken/1}, and \textit{best\_seller/1}. The total number of fact predicates in the menu knowledge base reaches 1220.

\smallskip\noindent\textbf{System Updating:}
In AutoManager, the manager and service bot are required to maintain a list of the current state. However, the s(CASP) goal-directed ASP system does not support automatic state updates. Therefore, we constructed a general updating system for the bots.

When a new predicate comes, we add a prefix ``new\_'' to the predicate, and put it in the waiting list. For example, a newly received \textit{runout/1} will be marked as \textit{new\_runout/1}. A consistency-checking template will check if the new predicate is consistent with the previous state predicates and generate the updated predicates accordingly. For example, if the ingredient has been restored, the rule below will remove the previous \textit{runout/1} predicate, and the \textit{new\_restore} predicate will not be added to the state. 

\begin{lstlisting}[style=MySCASP]
updated_runout(X) :- runout(X), not new_restore(X).
\end{lstlisting}

\noindent Then, the reasoning engine uses the predicates in both the state and the waiting list for the next action generation. Finally, all available predicates will be added to the state without the prefix.
In the service bot, the case is more complicated. For example, if the customer wants to upgrade a dish he or she did not order previously, or the dish is no longer available, the assistant bot will reject adding it to the state by following the rules.

\begin{lstlisting}[style=MySCASP]
updated_update(Dish, Operation, Option, I) :- 
    new_update(Dish, Operation, Option), all_order(Dish, N),
    dish(Dish), available_operation(Dish, Operation, I), 
    not unavailable(Dish). 
all_order(Dish, N) :- updated_order(Dish, N).
all_order(Dish, N) :- new_order(Dish, N).
\end{lstlisting}

\noindent The confirmation predicate \textit{confirm(unavailable, Reason)} will be generated simultaneously to explain the unavailability to the user. Since one can order a dish multiple times, we assign a number $I$ for each dish. In the previous example, this is done by the predicate \textit{available\_operation(Dish, Operation, I)}. Some operations like ``no onion'' or ``large size'' can only be claimed once for a dish, and this consistency check is also done by the \textit{available\_operation/3} predicate.

\smallskip\noindent\textbf{Leveraging LLMs to Correct Names:}
In the assistant bot, the knowledge extraction quality highly depends on the correctness of the extracted dish or combo name. However, the user input can greatly affect the LLM to find the correct name. If the customer misspelled it (we assume that the customers type their input), the LLM parser will follow the customer's mistake, resulting in a mismatch. In this case, we ask the LLM to find the dish name that best matches it. For each dish and combo, we use an LLM to transfer its name to embeddings, stored in the knowledge base. Once the dish name in the extracted predicate does not match any existing name exactly, we send it to the same LLM querying for the embeddings and use the cosine similarity to calculate the most similar name. This dish will replace the originally extracted one in the later reasoning. This is also applied in the administrator bot for ingredient name correction.

\section{Conversation Samples}
Below is an example of the dialogue generated from the dual-system bots for illustration. First we illustrate the administrator bot, then the assistant bot.

\begin{description}[align=right,labelindent=!,leftmargin=2cm]
\tt \small
\item[ManagerBot:] Hey there, can I assist you with anything?
\item[Employee:] We have no more slow-roasted chicken.
\end{description}
\vspace{-0.08in}
\begin{description}
\tt \small
\item[\color{gray}extracted semantics]
\end{description}
\vspace{-0.08in}
\begin{lstlisting}[style=MySCASP]
runout(Slow-Roasted Chicken).
\end{lstlisting}
\vspace{-0.08in}
\begin{description}
\tt \small
\item[\color{gray}next action]
\end{description}
\vspace{-0.08in}
\begin{lstlisting}[style=MySCASP]
confirm(runout, Slow-Roasted Chicken).
\end{lstlisting}
\vspace{-0.1in}
\begin{description}[align=right,labelindent=!,leftmargin=2cm]
\tt \small
\item[ManagerBot:] Understood, we're running short on Slow-Roasted Chicken.
\item[Employee:] The tomatoes are out of stock now.
\end{description}
\vspace{-0.08in}
\begin{description}
\tt \small
\item[\color{gray}extracted semantics]
\end{description}
\vspace{-0.08in}
\begin{lstlisting}[style=MySCASP]
runout(Tomatoes).
\end{lstlisting}
\vspace{-0.08in}
\begin{description}
\tt \small
\item[\color{gray}next action]
\end{description}
\vspace{-0.08in}
\begin{lstlisting}[style=MySCASP]
confirm(runout, Tomatoes).
\end{lstlisting}
\vspace{-0.1in}
\begin{description}[align=right,labelindent=!,leftmargin=2cm]
\tt \small
\item[ManagerBot:] Understood. We're running short on tomatoes.
\end{description}

\noindent\textbf{\color{red}Then the service bot took a human customer's order.}

\begin{description}[align=right,labelindent=!,leftmargin=1.8cm]
\tt \small
\item[ServiceBot:] Greetings, how may I assist you?
\item[Customer:] Hi, can I have a soft chicken taco?
\end{description}
\vspace{-0.08in}
\begin{description}
\tt \small
\item[\color{gray}extracted semantics]
\end{description}
\vspace{-0.08in}
\begin{lstlisting}[style=MySCASP]
order(Cantina Chicken Soft Taco, 1).
\end{lstlisting}
\vspace{-0.08in}
\begin{description}
\tt \small
\item[\color{gray}next action]
\end{description}
\vspace{-0.08in}
\begin{lstlisting}[style=MySCASP]
confirm(unavailable,[unavailable(Cantina Chicken Soft 
    Taco,runout(Slow-Roasted Chicken))]). else.
\end{lstlisting}
\vspace{-0.1in}

\begin{description}[align=right,labelindent=!,leftmargin=1.8cm]
\tt \small
\item[ServiceBot:] Apologies, but we're currently unable to provide the Cantina Chicken Soft
Taco as we have run out of Slow-Roasted Chicken. What additional items
would you like to include in your order?
\item[Customer:] Are there any popular tacos you recommend?
\end{description}
\vspace{-0.08in}
\begin{description}
\tt \small
\item[\color{gray}extracted semantics]
\end{description}
\vspace{-0.08in}
\begin{lstlisting}[style=MySCASP]
need_recommend(taco, category).
\end{lstlisting}
\vspace{-0.08in}
\begin{description}
\tt \small
\item[\color{gray}next action]
\end{description}
\vspace{-0.08in}
\begin{lstlisting}[style=MySCASP]
recommend(category, Soft Taco).
\end{lstlisting}
\vspace{-0.1in}

\begin{description}[align=right,labelindent=!,leftmargin=1.8cm]
\tt \small
\item[ServiceBot:] We suggest you sample our Soft Taco from the menu selection.
\item[Customer:] Great, I'd have two.
\end{description}
\vspace{-0.08in}
\begin{description}
\tt \small
\item[\color{gray}extracted semantics]
\end{description}
\vspace{-0.08in}
\begin{lstlisting}[style=MySCASP]
order(Soft Taco, 2).
\end{lstlisting}
\vspace{-0.08in}
\begin{description}
\tt \small
\item[\color{gray}next action]
\end{description}
\vspace{-0.08in}
\begin{lstlisting}[style=MySCASP]
confirm(order,Soft Taco). else.
\end{lstlisting}
\vspace{-0.1in}

\begin{description}[align=right,labelindent=!,leftmargin=1.8cm]
\tt \small
\item[ServiceBot:] You've selected a Soft Taco. Is there anything else you'd like to include
in your order?
\item[Customer:] One Pepsi, please.
\end{description}
\vspace{-0.08in}
\begin{description}
\tt \small
\item[\color{gray}extracted semantics]
\end{description}
\vspace{-0.08in}
\begin{lstlisting}[style=MySCASP]
order(Pepsi, 1).
\end{lstlisting}
\vspace{-0.08in}
\begin{description}
\tt \small
\item[\color{gray}next action]
\end{description}
\vspace{-0.08in}
\begin{lstlisting}[style=MySCASP]
confirm(order,Pepsi). else.
\end{lstlisting}
\vspace{-0.1in}

\begin{description}[align=right,labelindent=!,leftmargin=1.8cm]
\tt \small
\item[ServiceBot:] Understood, you'd like a Pepsi. What other dishes would you like to add to 
your order?
\item[Customer:] No thanks, that's all I need.
\end{description}
\vspace{-0.08in}
\begin{description}
\tt \small
\item[\color{gray}extracted semantics]
\end{description}
\vspace{-0.08in}
\begin{lstlisting}[style=MySCASP]
completed.
\end{lstlisting}
\vspace{-0.08in}
\begin{description}
\tt \small
\item[\color{gray}next action]
\end{description}
\vspace{-0.08in}
\begin{lstlisting}[style=MySCASP]
confirm(complete). ask([none,Soft Taco], add ingredients
    or sauces).
\end{lstlisting}
\vspace{-0.1in}

\begin{description}[align=right,labelindent=!,leftmargin=1.8cm]
\tt \small
\item[ServiceBot:] Thank you for placing your order. What additional toppings or sauces would
you prefer on your Soft Taco?
\item[Customer:] Can I have some tomatoes?
\end{description}
\vspace{-0.08in}
\begin{description}
\tt \small
\item[\color{gray}extracted semantics]
\end{description}
\vspace{-0.08in}
\begin{lstlisting}[style=MySCASP]
update(Soft Taco, add, Tomatoes).
\end{lstlisting}
\vspace{-0.08in}
\begin{description}
\tt \small
\item[\color{gray}next action]
\end{description}
\vspace{-0.08in}
\begin{lstlisting}[style=MySCASP]
confirm(unavailable,[unavailable(Tomatoes,runout(none))]).
    ask([none,Soft Taco], add ingredients or sauces).
\end{lstlisting}
\vspace{-0.1in}

\begin{description}[align=right,labelindent=!,leftmargin=1.8cm]
\tt \small
\item[ServiceBot:] We've run out of tomatoes at the moment. What other sauces or ingredients 
would you like in your Soft Taco?
\item[Customer:] What do people usually add?
\end{description}
\vspace{-0.08in}
\begin{description}
\tt \small
\item[\color{gray}extracted semantics]
\end{description}
\vspace{-0.08in}
\begin{lstlisting}[style=MySCASP]
need_recommend(Soft Taco, upgrade).
\end{lstlisting}
\vspace{-0.08in}
\begin{description}
\tt \small
\item[\color{gray}next action]
\end{description}
\vspace{-0.08in}
\begin{lstlisting}[style=MySCASP]
recommend(upgrade, Beans).
\end{lstlisting}
\vspace{-0.1in}

\begin{description}[align=right,labelindent=!,leftmargin=1.8cm]
\tt \small
\item[ServiceBot:] We suggest enhancing your meal by adding beans.
\item[Customer:] Great! Then add them to both tacos.
\end{description}
\vspace{-0.08in}
\begin{description}
\tt \small
\item[\color{gray}extracted semantics]
\end{description}
\vspace{-0.08in}
\begin{lstlisting}[style=MySCASP]
update(Soft Taco, add, Beans). update(Soft Taco,
    add, Beans).
\end{lstlisting}
\vspace{-0.08in}
\begin{description}
\tt \small
\item[\color{gray}next action]
\end{description}
\vspace{-0.08in}
\begin{lstlisting}[style=MySCASP]
confirm(add,Beans). ask([none,Soft Taco], make it fresco).
\end{lstlisting}
\vspace{-0.1in}

\begin{description}[align=right,labelindent=!,leftmargin=1.8cm]
\tt \small
\item[ServiceBot:] We've included beans in your order. Would you like your Soft Taco to be
prepared fresco style?
\item[Customer:] No thanks.
\end{description}
\vspace{-0.08in}
\begin{description}
\tt \small
\item[\color{gray}extracted semantics]
\end{description}
\vspace{-0.08in}
\begin{lstlisting}[style=MySCASP]
update(Soft Taco, fresco, no).
\end{lstlisting}
\vspace{-0.08in}
\begin{description}
\tt \small
\item[\color{gray}next action]
\end{description}
\vspace{-0.08in}
\begin{lstlisting}[style=MySCASP]
confirm(fresco,no). ask([none,Pepsi], choose size).
\end{lstlisting}
\vspace{-0.1in}

\begin{description}[align=right,labelindent=!,leftmargin=1.8cm]
\tt \small
\item[ServiceBot:] I understand that you prefer not to have it fresco. Could you please tell me 
the size you prefer for your Pepsi?
\item[Customer:] I'd like the regular size.
\end{description}
\vspace{-0.08in}
\begin{description}
\tt \small
\item[\color{gray}extracted semantics]
\end{description}
\vspace{-0.08in}
\begin{lstlisting}[style=MySCASP]
update(Pepsi, size, regular).
\end{lstlisting}
\vspace{-0.08in}
\begin{description}
\tt \small
\item[\color{gray}next action]
\end{description}
\vspace{-0.08in}
\begin{lstlisting}[style=MySCASP]
confirm(none,complete). order(Pepsi).  update(size,regular).  
order(Soft Taco). update(fresco,no).  update(add,Beans).  
order(Soft Taco). update(fresco,no).  update(add,Beans). 
price(7.57).
\end{lstlisting}
\vspace{-0.1in}

\begin{description}[align=right,labelindent=!,leftmargin=1.8cm]
\tt \small
\item[ServiceBot:] Thank you for your order. You've chosen a regular size Pepsi, and two Soft
Tacos without fresco, but with beans. The total comes to \$7.57.
\item[Customer:] Here it is. Thank you! Bye!
\end{description}
\vspace{-0.08in}
\begin{description}
\tt \small
\item[\color{gray}extracted semantics]
\end{description}
\vspace{-0.08in}
\begin{lstlisting}[style=MySCASP]
quit.
\end{lstlisting}
\vspace{-0.08in}
\begin{description}
\tt \small
\item[\color{gray}next action]
\end{description}
\vspace{-0.08in}
\begin{lstlisting}[style=MySCASP]
confirm(quit, none). quit.
\end{lstlisting}
\vspace{-0.1in}
\begin{description}[align=right,labelindent=!,leftmargin=1.8cm]
\tt \small
\item[ServiceBot:] Thank you for choosing Taco Bell. Have a great day! Goodbye!
\end{description}

\noindent\textbf{\color{red}The customer session ends.}

\begin{description}[align=right,labelindent=!,leftmargin=2cm]
\tt \small
\item[Employee:] Let's end today's work.
\end{description}
\vspace{-0.08in}
\begin{description}
\tt \small
\item[\color{gray}extracted semantics]
\end{description}
\vspace{-0.08in}
\begin{lstlisting}[style=MySCASP]
quit.
\end{lstlisting}
\vspace{-0.15in}
\begin{description}
\tt \small
\item[\color{gray}next action]
\end{description}
\vspace{-0.08in}
\begin{lstlisting}[style=MySCASP]
confirm(quit, none). quit.
\end{lstlisting}
\vspace{-0.08in}
\begin{description}[align=right,labelindent=!,leftmargin=2cm]
\tt \small
\item[ManagerBot:] Alright, I got it. You want me to quit for the day. Bye!
\end{description}

\section{Evaluation}
We report on the efficiency of our system, along with human evaluation of dialog transcripts generated by AutoManager with 
real-world Taco Bell Drive-Thru AI System.  

\begin{table}[tb]
    \centering
    \resizebox{0.9\linewidth}{!}{
    \begin{tabular}{lccc}
    \hline
         Call Option & Manager & Service (1) & Service (10)\\
    \hline
        s(CASP) Reasoning (for one call) & 9.9ms & 115.4ms & 886.5ms\\

        Total (for each round) &  2637.5ms & 3384.1 & 4215.7ms\\
    \hline
    \end{tabular}}
    \caption{Time cost for s(CASP) reasoning call and total response generation in the manager bot and the customer service bot of AutoManager. Here ``1'' means the service bot is called with only one requirement predicate in the list, and ``10'' means 10 predicates are captured in the state list.}
    \label{tab:manager_time}
\end{table}

\smallskip\noindent\textbf{Efficiency Evaluation}.
We experimented on the processing time of the manager and service bot for each round of reply generation in Table \ref{tab:manager_time}. The average time consumption for 50 rounds is around 2.6 seconds for the administrator bot and 4.2 seconds for the assistant bot for 10 user requirements, which is an acceptable waiting time for serving sessions. The result also shows that the average time consumption for ASP reasoning is about 9.9 milliseconds for the administrator bot. For the assistant bot, the time consumption varies by the size of the user's requirements. As the number of user requirements increased, it took longer for s(CASP) to reason. However, when the requirement number comes to 10, which is enough for an ordinary customer, the time spent on s(CASP) is only 886.5 milliseconds, which is a reasonable computation time. In addition, the time cost of the reasoning step is much shorter than the total time consumption, which also indicates that most of the time is spent calling GPT-4. We believe that LLMs with faster response times will emerge shortly, and the overall execution time will be further shortened.

\begin{table}[ht]
    \centering
    \resizebox{0.9\linewidth}{!}{
    \begin{tabular}{lcc}
    \hline
       & AutoManager (ours) & Taco Bell \\
      \hline
      Understanding  & \textbf{9.18} & 8.04 \\
      Truthfulness & \textbf{9.50} & 9.00 \\
      Coherency & \textbf{9.16} & 8.60 \\
      Fluency & \textbf{8.96} & \textbf{8.96}\\
      Task Ease & \textbf{8.88} & 7.72\\
      Expected Behavior & \textbf{9.60} & 8.20 \\
      Total Satisfaction & \textbf{9.00} & 8.12 \\\hline
    \end{tabular}}
    \caption{Human Evaluation Results comparing our AutoManager with the real-world Taco Bell Drive-Thru AI system. We tested on five conversations with five human evaluators.}
    \label{tab:human}
\end{table}

\smallskip\noindent\textbf{Human Evaluation}.
We also manually evaluated the conversation quality of our AutoManager and compared its results with the real-world Taco Bell drive-thru AI order-taking system. Note that the latter is a custom implementation watched over by a human. Since we do not have access to the Taco Bell drive-thru AI system, we collected the reported conversations through the Internet. We went through the videos posted online that include the entire conversation with Taco Bell AI and extracted transcripts from them. Then, we applied these conversations in our AutoManager system following the same requirements for the orders. As a result, we generated five pairs of conversations, where each pair contains one conversation from the Taco Bell drive-thru AI and one from AutoManager. We asked five well-educated graders to grade these five conversation pairs on seven metrics, following the human evaluation on the AutoConcierge by \cite{autoconcierge}: \textbf{Understanding}, \textbf{Truthfulness}, \textbf{Coherency}, \textbf{Fluency}, \textbf{Task Ease}, \textbf{Expected Behavior}, and \textbf{Total Satisfactory}. The result can be found in Table \ref{tab:human}. The result shows that AutoManager leads dominantly in understanding, task ease, and expected behavior, indicating that the core advantage of AutoManager over Taco Bell AI is its reliability and human-like behavior.

\section{Conclusion and Future Work}
This paper provides a paradigm for building a dual-chatbot collaboration system using an LLM as a parser and an ASP system as the reasoner. In this paradigm, each chatbot is built using the STAR framework, where an LLM extracts the knowledge predicates from the user input, and then an ASP reasoning engine, augmented with commonsense knowledge, reasons over these extracted predicates to check their consistency and generate the next action output as predicates. The output predicates are converted to natural language sentences by using the LLM yet again. The two bots share a knowledge base, temporary information lists, and the collaborative rule sets, and coordinate with each other by ``leveraging external information for internal tasks'' mechanism for high security and efficiency. AutoManager has been evaluated manually, and it demonstrates greater reliability than the real-world Taco Bell drive-thru AI system.
%
Our future work is to explore a paradigm where a larger number of collaborative chatbots work with humans.
For example, 
%
in a hospital, a counselor is responsible for helping the user to register, a specialist in various disciplines diagnoses the user's condition, a pharmacist provides the user with prescribed medication, and so on. These agents share the user's medical data, 
the status of the hospital equipment, instruments, medicines,
etc. 
Such chatbots may pose different challenges, since as the shared knowledge grows exponentially, the time to find and manage it increases exponentially. Therefore, it may be necessary to introduce a shared knowledge manager that filters out and categorizes the knowledge that different agents may use in advance.

\bibliographystyle{IEEEtran}
\bibliography{generic}

\end{document}